\documentclass[sigconf,natbib=true]{acmart}

\usepackage{bm}
\usepackage{amsmath}
\usepackage{balance}

\AtBeginDocument{%
  \providecommand\BibTeX{{%
    \normalfont B\kern-0.5em{\scshape i\kern-0.25em b}\kern-0.8em\TeX}}}

\setcopyright{acmcopyright}
\copyrightyear{2021}
\acmYear{2021}
\acmDOI{}

\acmConference[CIKM '21]{Proceedings of the 30th ACM International Conference on Information and Knowledge Management}{November 1--5, 2021}{Virtual Event, QLD, Australia}\acmBooktitle{Proceedings of the 30th ACM International Conference on Information and Knowledge Management (CIKM '21), November 1--5, 2021, Virtual Event, QLD, Australia}
\acmPrice{15.00}
\acmISBN{}

\begin{document}

\title{Automated Hyperparameter Optimization Challenge at CIKM 2021 AnalyticCup}

\author{Huaijun Jiang}
\authornote{Equal contribution.}
\email{jianghuaijun@pku.edu.cn}
\affiliation{%
  \institution{Peking University}
  \city{Beijing}
  \country{China}
}

\author{Yu Shen}
\authornotemark[1]
\email{shenyu@pku.edu.cn}
\affiliation{%
  \institution{Peking University}
  \city{Beijing}
  \country{China}
}

\author{Yang Li}
\authornotemark[1]
\email{liyang.cs@pku.edu.cn}
\affiliation{%
  \institution{Peking University}
  \city{Beijing}
  \country{China}
}

\renewcommand{\shortauthors}{Huaijun Jiang, et al.}

\begin{abstract}
In this paper, we describe our method for tackling the automated hyperparameter optimization challenge in QQ Browser 2021 AI Algorithm Competiton (ACM CIKM 2021 AnalyticCup Track 2). The competition organizers provide anonymized realistic industrial tasks and datasets for black-box optimization. Based on our open-sourced package OpenBox, we adopt the Bayesian optimization framework for configuration sampling and a heuristic early stopping strategy. We won first place in both the preliminary and final contests with the results of 0.938291 and 0.918753, respectively.

\end{abstract}

\maketitle

\section{Introduction}
The automated hyperparameter optimization challenge of ACM CIKM 2021 AnalyticCup is an annual contest that evaluates automated hyperparameter optimization algorithms on 30 anonymized realistic tasks. 
The data is constructed on industrial recommendation scenarios, which aims at improving the practical performance of machine learning models and strategies and reducing human effort in choosing hyperparameters based on experts' experience.

In general, the contest can be viewed as a black-box optimization problem, in which the performance of a set of hyperparameters can only be obtained via a specific function $f$ without analytical forms. The problem can be formulated as,
\begin{equation}
    \mathop{\arg\min}_{\bm{x} \in \mathcal{X}}f(\bm{x}),
    \label{eq:problem}
\end{equation}
where $\mathcal{X}$ is the hyperparameter space and $\bm{x}$ is a choice from the space. The goal is find the best choice that minimizes the given function $f$.

Bayesian optimization (BO)~\cite{snoek2012practical,hutter2011sequential,bergstra2011algorithms} is a typical framework for solving black-box optimization tasks shown in Equation ~\ref{eq:problem}. 
It contains two core components that allow for improved exploration over the search space. The first component is the probabilistic surrogate model that approximates $f({\bm x})$ based on historical evaluations and outputs uncertainty estimates to guide exploration. The other component is the acquisition function that measures the utility by trading off exploration and exploitation.

In this contest, we elaborately select each component for Bayesian optimization and design a heuristic early stopping strategy.
The solution is implemented based on OpenBox~\cite{li2021openbox}, a generalized service for black-box optimization. In the following sections, we first give a brief introduction to the Bayesian optimization framework, and then introduce our solutions for the preliminary and final contest.

\section{Bayesian Optimization Framework}
In this section, we briefly introduce the framework of Bayesian Optimization (BO). 
We denote a configuration $\bm x$ as a choice of hyperparameters from their value ranges and observations $D$ as the collection of configurations and their corresponding performance.
Since evaluating the objective function $f$ for a
given configuration $\bm{x}$ is very expensive, BO approximates $f$ using a probabilistic surrogate model $M:p(f|D)$ that is much cheaper to evaluate.
Given a configuration $\bm{x}$, the surrogate model $M$ outputs the posterior predictive distribution at $\bm{x}$, that is,
$f(\bm{x}) \sim \mathcal{N}(\mu_{M}(\bm{x}), \sigma^2_{M}(\bm{x}))$.
In the $n^{th}$ iteration, BO methods iterate the following three steps: 
1) use the surrogate model $M$ to select a configuration that maximizes the acquisition function $\bm{x}_{n}=\arg\max_{\bm{x} \in \mathcal{X}}a(\bm{x}; M)$, where the acquisition function is used to balance the exploration and exploitation; 
2) evaluate the configuration $\bm{x}_{n}$ to get its performance $y_{n}=f(\bm{x}_{n})+\epsilon$ with $\epsilon \sim \mathcal{N}(0, \sigma^2)$; 
3) add this measurement $(\bm{x}_{n}, y_{n})$ to observations $D = \{(\bm{x}_1, y_1),...,(\bm{x}_{n-1}, y_{n-1})\}$, and refit the surrogate model on the augmented $D$.

\section{Preliminary Contest}
The preliminary contest is a typical black-box optimization problem. In each iteration, the optimizer should suggest 5 configurations from a given search space in a synchronous manner. Rewards will return afterward for the next round of suggestion. 
The objective is to maximize the reward within 100 hyperparameter evaluations.
We will introduce our method for the preliminary contest in detail in this section.

\subsection{Configuration Space Definition}
We use the package ConfigSpace~\footnote{https://github.com/automl/ConfigSpace} to define the configuration (hyperparameter) space. 
Since the choices of all hyperparameters are of approximately equal distance, we index the valid values of hyperparameters by using the class \texttt{UniformIntegerHyperparameter}. The value ranges are scaled to $[0, 1]$ when fitting the surrogate.

\subsection{Initial Design}
To start up the Bayesian optimization, we design a greedy algorithm to generate initial configurations. The algorithm selects configurations sequentially from randomly sampled candidates until a preset size is reached. In each round, it selects the furthest configuration from the existing selected ones. In online testing, this method performs better than naive random design. The method is provided in OpenBox as the ``random\_explore\_first'' initial design. We set the number of initial configurations to 10.

\subsection{Surrogate Model}
Since the number of hyperparameters is relatively small, and all hyperparameters are numerical, we adopt the Gaussian Process~\cite{snoek2012practical} as the surrogate.
Given a new configuration, the Gaussian process outputs a normal distribution, whose mean and variance are formulated as follows,
\begin{equation}
\begin{aligned}
\mu_{M}(\bm{x}) &= K_* K^{-1} y,\\
\sigma^2_{M}(\bm{x}) &= K_{**} - K_*K^{-1}K_*^T,
\end{aligned}
\end{equation}
where $K$, $K_*$, $K_{**}$ are computed according to the kernel function,
\begin{gather}
    K =
    \begin{bmatrix}
    k(\bm{x}_1,\bm{x}_1) & k(\bm{x}_1,\bm{x}_2) & \cdots & k(\bm{x}_1,\bm{x}_n) \\
    k(\bm{x}_2,\bm{x}_1) & k(\bm{x}_2,\bm{x}_2) & \cdots & k(\bm{x}_2,\bm{x}_n) \\
    \vdots & \vdots & \ddots & \vdots \\
    k(\bm{x}_n,\bm{x}_1) & k(\bm{x}_n,\bm{x}_2) & \cdots & k(\bm{x}_n,\bm{x}_n)
    \end{bmatrix}, \notag \\
    K_* =
    \begin{bmatrix}
    k(\bm{x}_*,\bm{x}_1) & k(\bm{x}_*,\bm{x}_2) & \cdots & k(\bm{x}_*,\bm{x}_n) \\
    \end{bmatrix},\\
    K_{**} = k(\bm{x}_*,\bm{x}_*). \notag 
\end{gather}

Compared with other surrogates, such as Probabilistic Random Forest~\cite{hutter2011sequential}, Tree Parzen Estimator~\cite{bergstra2011algorithms}, etc., Gaussian Process is suitable to apply on continuous output space with numerical inputs.
In the Gaussian process, we use the Mat\'ern52 kernel and use the default settings for the kernel.

\subsection{Acquisition Function and Optimization}
We adopt the Expected Improvement (EI) function ~\cite{jones1998efficient} as the acquisition function, which can be formulated as,
\begin{equation}
    \begin{small}
        EI(\bm{x}; D)=\int_{-\infty}^{\infty} \max(y_{min}-y, 0)p_{D}(y|\bm{x})dy,
        \label{eq:ei}
    \end{small}
\end{equation}
where $y_{min}$ is the best performance observed so far. Given observations $D$, the EI function computes the expectation of improving the current best performance.
To maximize the EI function, we further apply the L-BFGS-B algorithm~\cite{Zhu94l-bfgs-b-}, which is a popular algorithm for parameter estimation in machine learning.  
While the results of L-BFGS-B largely depend on the choice of initial points, we adopt the following two sampling strategies: 
1) Randomly sample a large number of configurations via the Monte Carlo sampling from the entire space;
2) Sample a few points from the best-observed configurations by changing the value of only one hyperparameter.
We combine the configurations sampled by the two strategies, and select 10 configurations with the largest EI value as the start points for L-BFGS-B. 
This method is provided in OpenBox as the ``random\_scipy'' acquisition optimizer.

\subsection{Other Techniques}
To suggest multiple configurations in one iteration, we adopt a simple method. We restart the optimization of the acquisition function and obtain one configuration each time until a desired number of suggestions is reached.
To increase the exploratory of Bayesian optimization and ensure the convergence of the algorithm, we set the probability of suggesting random configuration to 10\%.

\section{Final Contest}
The final contest is a more challenging optimization problem. 
The evaluation process of each configuration is divided into 14 iterations, with partially evaluated rewards in the first 13 iterations and fully evaluated reward in the last iteration reported. 
The 95\% confidence interval information for the reward is additionally provided. 
The optimizer should decide whether to early stop the evaluation process of a configuration to speed up optimization. 
The objective is kept the same, but only the configurations evaluated for 14 iterations will be recorded as valid ones.
The time budget is reduced to half of the preliminary stage, which means at most 50 configurations could be fully evaluated.
In this section, we will present our solution in detail.

\subsection{Data Analysis}
To design a promising strategy for the final contest, we first look into the opened local datasets and find some interesting properties:

\begin{itemize}
\item Given any confidence interval reported in the first 13 iterations of a configuration, the probability that the final reward in the 14-th iteration of the same configuration is located in the interval is perfectly 95\%.
\item Given any reward in the first 13 iterations of a configuration, the probability that the final reward in the 14-th iteration of the same configuration is greater than the given reward is perfectly 50\%.
\item For any configuration, during the 14 iterations of evaluation, the confidence interval is reduced, however, the reward values show a random trend.
\end{itemize}

We also explore the order relation between configurations. Given any two configurations and their rewards of any iteration in the first 13 iterations, we compare the order of the given rewards with the order of their final rewards in the 14-th iteration and find that:

\begin{itemize}
\item Among all configurations, the consistency of the reward orders between partial evaluation and full evaluation is greater than 95\%.
\item Among configurations with top 1\% final reward in search space, the consistency of order of rewards between partial evaluation and full evaluation is about 50\% to 70\%.
\end{itemize}

\begin{figure}
    \centering
    \scalebox{0.4}{\includegraphics{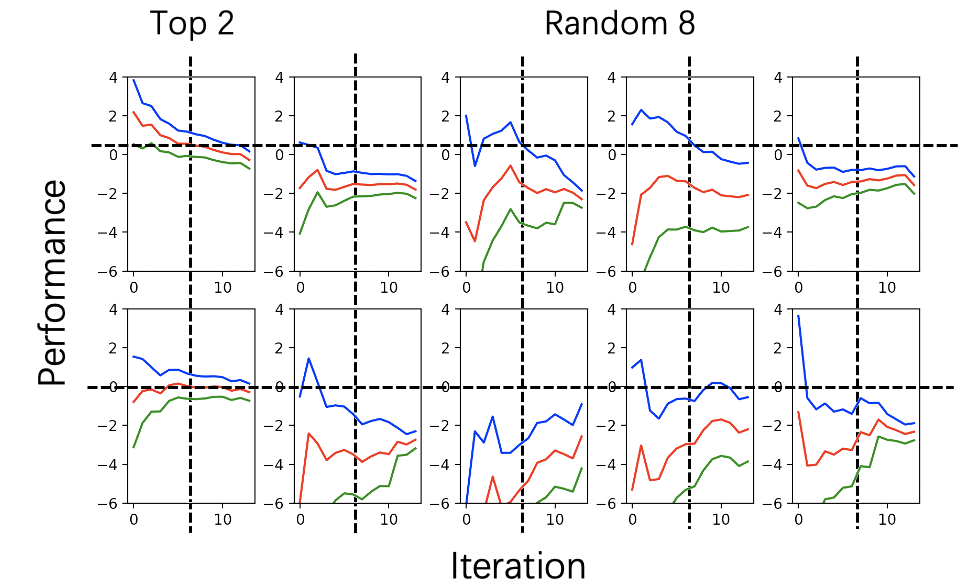}}
    \vspace{-3mm}
    \caption{The upper bound (blue), median (red), lower bound (green) of the best 2 configurations and 8 random configurations.}
    \label{fig:observation}
    \vspace{-5mm}
\end{figure}

As shown in Figure~\ref{fig:observation}, we plot the performance of the best 2 configurations and 8 random configurations over iteration. The median performance of the top-2 configurations outperforms the random ones at the 7-th iteration, which indicates we can pick out those badly-performing configurations in early iterations. However, by the same way, it is hard to identify the best configuration among well-performing configurations because of non-negligible noise in partial evaluations.

\subsection{Early Stopping}
Based on the observations, we consider a heuristic early stopping strategy. 
However, there is a trade-off on when to start early stopping. 
On one hand, early stopping speeds up evaluation and explores more configurations. 
On the other hand, full evaluation provides accurate reward information, which benefits the surrogate in Bayesian optimization. 
In our final implementation, we begin early stopping after 40 configurations are fully evaluated. 
Our early stopping strategy is inspired by ASHA~\cite{li2020system_asha}, in which we compare the reward at the 7-th iteration of the running configuration with all rewards observed in the same iteration. 
If the reward is not in the top half, the evaluation is stopped and a new configuration should be suggested.

\subsection{Value Imputation}
Since the partial results involve non-negligible noises, we apply BO only on those fully evaluated configurations.
However, if we early stop a configuration, the BO framework will not receive its feedback and may suggest a similar configuration in the next round.
As a result, we assume that the stopped configuration is not a good one, and then we impute its final performance with the median of all the observed results.

\section{Conclusion}
In this technical report, we presented our winning solution for the automated hyperparameter optimization challenge at ACM CIKM 2021 AnalyticCup. 
We adopted the Bayesian optimization in the preliminary contest and added a heuristic early stopping strategy in the final contest.
Our implementation for the preliminary contest has been integrated into OpenBox. Please refer to the Github repository~\footnote{https://github.com/PKU-DAIR/open-box} for more details. 




\bibliographystyle{ACM-Reference-Format}
\bibliography{sample-base}


\begin{thebibliography}{7}


\ifx \showCODEN    \undefined \def \showCODEN     #1{\unskip}     \fi
\ifx \showDOI      \undefined \def \showDOI       #1{#1}\fi
\ifx \showISBNx    \undefined \def \showISBNx     #1{\unskip}     \fi
\ifx \showISBNxiii \undefined \def \showISBNxiii  #1{\unskip}     \fi
\ifx \showISSN     \undefined \def \showISSN      #1{\unskip}     \fi
\ifx \showLCCN     \undefined \def \showLCCN      #1{\unskip}     \fi
\ifx \shownote     \undefined \def \shownote      #1{#1}          \fi
\ifx \showarticletitle \undefined \def \showarticletitle #1{#1}   \fi
\ifx \showURL      \undefined \def \showURL       {\relax}        \fi
\providecommand\bibfield[2]{#2}
\providecommand\bibinfo[2]{#2}
\providecommand\natexlab[1]{#1}
\providecommand\showeprint[2][]{arXiv:#2}

\bibitem[\protect\citeauthoryear{Bergstra, Bardenet, Bengio, and
  K{\'e}gl}{Bergstra et~al\mbox{.}}{2011}]%
        {bergstra2011algorithms}
\bibfield{author}{\bibinfo{person}{James~S Bergstra}, \bibinfo{person}{R{\'e}mi
  Bardenet}, \bibinfo{person}{Yoshua Bengio}, {and} \bibinfo{person}{Bal{\'a}zs
  K{\'e}gl}.} \bibinfo{year}{2011}\natexlab{}.
\newblock \showarticletitle{Algorithms for hyper-parameter optimization}. In
  \bibinfo{booktitle}{\emph{Advances in neural information processing
  systems}}. \bibinfo{pages}{2546--2554}.
\newblock


\bibitem[\protect\citeauthoryear{Hutter, Hoos, and Leyton-Brown}{Hutter
  et~al\mbox{.}}{2011}]%
        {hutter2011sequential}
\bibfield{author}{\bibinfo{person}{Frank Hutter}, \bibinfo{person}{Holger~H
  Hoos}, {and} \bibinfo{person}{Kevin Leyton-Brown}.}
  \bibinfo{year}{2011}\natexlab{}.
\newblock \showarticletitle{Sequential model-based optimization for general
  algorithm configuration}. In \bibinfo{booktitle}{\emph{International
  Conference on Learning and Intelligent Optimization}}. Springer,
  \bibinfo{pages}{507--523}.
\newblock


\bibitem[\protect\citeauthoryear{Jones, Schonlau, and Welch}{Jones
  et~al\mbox{.}}{1998}]%
        {jones1998efficient}
\bibfield{author}{\bibinfo{person}{Donald~R Jones}, \bibinfo{person}{Matthias
  Schonlau}, {and} \bibinfo{person}{William~J Welch}.}
  \bibinfo{year}{1998}\natexlab{}.
\newblock \showarticletitle{Efficient global optimization of expensive
  black-box functions}.
\newblock \bibinfo{journal}{\emph{Journal of Global optimization}}
  \bibinfo{volume}{13}, \bibinfo{number}{4} (\bibinfo{year}{1998}),
  \bibinfo{pages}{455--492}.
\newblock


\bibitem[\protect\citeauthoryear{Li, Jamieson, Rostamizadeh, Gonina, Ben-tzur,
  Hardt, Recht, and Talwalkar}{Li et~al\mbox{.}}{2020}]%
        {li2020system_asha}
\bibfield{author}{\bibinfo{person}{Liam Li}, \bibinfo{person}{Kevin Jamieson},
  \bibinfo{person}{Afshin Rostamizadeh}, \bibinfo{person}{Ekaterina Gonina},
  \bibinfo{person}{Jonathan Ben-tzur}, \bibinfo{person}{Moritz Hardt},
  \bibinfo{person}{Benjamin Recht}, {and} \bibinfo{person}{Ameet Talwalkar}.}
  \bibinfo{year}{2020}\natexlab{}.
\newblock \showarticletitle{A System for Massively Parallel Hyperparameter
  Tuning}.
\newblock \bibinfo{journal}{\emph{Proceedings of Machine Learning and Systems}}
   \bibinfo{volume}{2} (\bibinfo{year}{2020}), \bibinfo{pages}{230--246}.
\newblock


\bibitem[\protect\citeauthoryear{Li, Shen, Zhang, Chen, Jiang, Liu, Jiang, Gao,
  Wu, Yang, Zhang, and Cui}{Li et~al\mbox{.}}{2021}]%
        {li2021openbox}
\bibfield{author}{\bibinfo{person}{Yang Li}, \bibinfo{person}{Yu Shen},
  \bibinfo{person}{Wentao Zhang}, \bibinfo{person}{Yuanwei Chen},
  \bibinfo{person}{Huaijun Jiang}, \bibinfo{person}{Mingchao Liu},
  \bibinfo{person}{Jiawei Jiang}, \bibinfo{person}{Jinyang Gao},
  \bibinfo{person}{Wentao Wu}, \bibinfo{person}{Zhi Yang}, \bibinfo{person}{Ce
  Zhang}, {and} \bibinfo{person}{Bin Cui}.} \bibinfo{year}{2021}\natexlab{}.
\newblock \showarticletitle{OpenBox: A Generalized Black-box Optimization
  Service}.
\newblock \bibinfo{journal}{\emph{Proceedings of the 27th ACM SIGKDD Conference
  on Knowledge Discovery \& Data Mining}} (\bibinfo{year}{2021}).
\newblock


\bibitem[\protect\citeauthoryear{Snoek, Larochelle, and Adams}{Snoek
  et~al\mbox{.}}{2012}]%
        {snoek2012practical}
\bibfield{author}{\bibinfo{person}{Jasper Snoek}, \bibinfo{person}{Hugo
  Larochelle}, {and} \bibinfo{person}{Ryan~P Adams}.}
  \bibinfo{year}{2012}\natexlab{}.
\newblock \showarticletitle{Practical bayesian optimization of machine learning
  algorithms}. In \bibinfo{booktitle}{\emph{Advances in neural information
  processing systems}}. \bibinfo{pages}{2951--2959}.
\newblock


\bibitem[\protect\citeauthoryear{Zhu, Byrd, Lu, and Nocedal}{Zhu
  et~al\mbox{.}}{1994}]%
        {Zhu94l-bfgs-b-}
\bibfield{author}{\bibinfo{person}{Ciyou Zhu}, \bibinfo{person}{Richard~H.
  Byrd}, \bibinfo{person}{Peihuang Lu}, {and} \bibinfo{person}{Jorge Nocedal}.}
  \bibinfo{year}{1994}\natexlab{}.
\newblock \bibinfo{booktitle}{\emph{L-BFGS-B - Fortran Subroutines for
  Large-Scale Bound Constrained Optimization}}.
\newblock \bibinfo{type}{{T}echnical {R}eport}. \bibinfo{institution}{ACM
  Trans. Math. Software}.
\newblock


\end{thebibliography}





\end{document}